\documentclass[conference]{IEEEtran}
\IEEEoverridecommandlockouts

\usepackage{cite}
\usepackage{amsmath,amssymb,amsfonts}
\usepackage{algorithmic}
\usepackage{graphicx}
\usepackage{textcomp}
\usepackage{xcolor}
\usepackage{subfigure}
\usepackage{booktabs}
\usepackage{multirow}
\usepackage{array}
\usepackage{hyperref}

\def\BibTeX{{\rm B\kern-.05em{\sc i\kern-.025em b}\kern-.08em
    T\kern-.1667em\lower.7ex\hbox{E}\kern-.125emX}}
\begin{document}

\title{A Semi-Supervised Learning Approach for Abnormal Event Prediction on Large Network Operation Time-Series Data}


\author{\IEEEauthorblockN{Yijun Lin, Yao-Yi Chiang} 
\IEEEauthorblockA{\textit{Department of Computer Science and Engineering} \\
\textit{University of Minnesota, Twin Cities} \\
\{lin00786, yaoyi\}@umn.edu}
}

\maketitle 

\begin{abstract}
Large network logs, recording multivariate time series generated from heterogeneous devices and sensors in a network, can reveal important information about abnormal activities, such as network intrusions and packet losses. Existing machine learning methods for anomaly detection on multiple multivariate time series typically assume that 1) infrequent behaviors beyond some inference threshold are anomalous for unsupervised models or 2) require a large set of labeled normal and abnormal sequences for supervised models. However, in practice, the reported abnormal events might be available but incomplete and sparse (i.e., much fewer than normal cases). This paper presents a novel semi-supervised approach, SNetAD, that takes advantage of the incomplete and imbalanced labels to effectively learn separable feature embeddings of network activities representing normal and abnormal events. Specifically, SNetAD first generates network representations by capturing relationships across time points and between network devices. Then SNetAD encourages the embeddings to form two clusters using contrastive center loss and improves the separability of the learned clusters using labeled and unlabeled samples in a semi-supervised manner. The experiments demonstrate that SNetAD significantly outperforms state-of-the-art approaches for abnormal event prediction on a large real-world network log.

\end{abstract}

\begin{IEEEkeywords}
multivariate time series, semi-supervised, network event prediction
\end{IEEEkeywords}

\section{Introduction}

\begin{figure}[htbp]
\centering
\includegraphics[width=\linewidth]{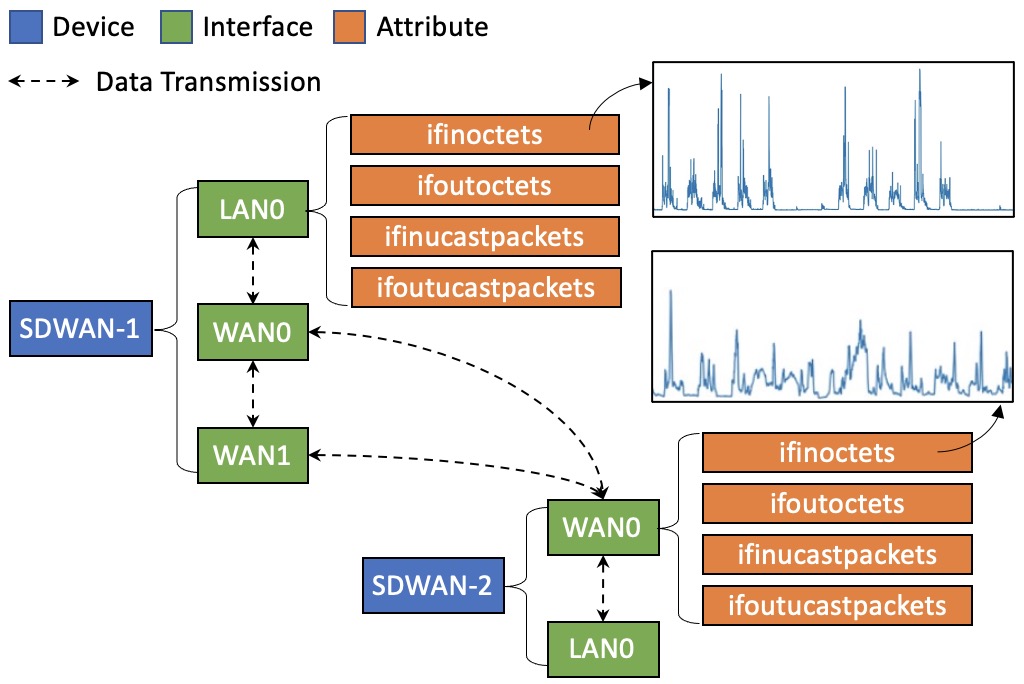}
\caption{An example network contains two SDWAN devices, and the devices have some LAN and WAN interfaces. The interfaces can behave differently, measured by four attributes over time. There exists some relationships between interfaces due to data transmission.}
\label{figure: network_def}
\end{figure}

Nowadays, many network management companies (e.g., Cisco, NTT Global Networks) collect large volumes of data logs to measure and monitor the status (e.g., healthy, overloaded, power outage) of devices (e.g., routers, switches) or interfaces (e.g., local area networks (LAN), wide area networks (WAN)) in a network. Figure~\ref{figure: network_def} shows an example network. A network usually consists of multiple devices, each with some interfaces, e.g., an SDWAN device has LAN and WAN interfaces. The network topology is usually unknown (e.g., not provided by users). There could be connections (e.g., data transmission) between interfaces via the linkage. Each interface is measured by several attributes, i.e., the number of in/out octets and in/out ucast packets, that describe network operations or activities as multivariate time series. Thus, these measurements for the entire network can serve as \textit{multiple multivariate time series}.\footnote{Here, ``multiple'' indicates multiple interfaces, each measured by several variables as ``multivariate'' over time.}

Effectively predicting abnormal events from network data logs is essential, which helps continuously monitor network states and raise alerts for potential incidents on time for network engineers to investigate and resolve. Traditional monitoring systems typically trigger a real-time notification when the network is experiencing some abnormal event (e.g., packet loss, intrusion) using pre-defined rules but cannot provide an early warning on a potential incident. In addition, a network 
can contain hundreds of heterogeneous devices from different vendors with various criteria (e.g., some rules and thresholds) indicating abnormal status, making it challenging to build an early warning system for abnormal events and enable the system to handle all kinds of networks.

Advanced data-driven methods (e.g., machine learning approaches) for anomaly detection on \textit{multiple multivariate time series} can automatically learn representations for normal (and abnormal) events from network operation data without domain knowledge (e.g., incident criteria) by jointly capturing (1) \textit{the temporal dependencies across time points} using recurrent neural networks (RNNs)~\cite{malhotra2016lstm, hundman2018detecting} and Transformers~\cite{cai2020traffic, tuli2022tranad}, and (2) \textit{the relationships between time series} using feature fusion~\cite{malhotra2016lstm, li2019mad}, graph convolution~\cite{kipf2016semi, chen2022deep}, and graph attention~\cite{zhou2022hybrid, fan2020anomalydae}. Then these learned representations serve to identify anomalies using classification or threshold strategies~\cite{schmidl2022anomaly}. There are supervised models~\cite{liu2015opprentice, shipmon2017time} that leverage labeled data to guide the process of learning normal and abnormal behaviors in the time series, but they usually require sufficient labeled samples. However, anomaly labeling requires high expert costs, and the labeling criteria could vary significantly among devices and interfaces, e.g., an abnormal event can last for several minutes, but only the beginning time point is marked as anomalous. Thus, many existing anomaly detection approaches are unsupervised that identify infrequent behaviors in multivariate time series as an anomaly, including reconstruction-based methods~\cite{malhotra2016lstm, park2018multimodal, su2019robust}, forecasting-based methods~\cite{laptev2015generic, hundman2018detecting, akoglu2010event}, and hybrid methods~\cite{zhao2020multivariate, chen2022deep}. However, in practice, some labeled data might be available to guide the learning process, e.g., manually issued event tickets\footnote{Some users might report incidents to the vendors.}, but typically (1)~\textit{incomplete}, only a portion of the network operations can be labeled based on the event tickets\footnote{This is because there could be a delay (e.g., 15 minutes or even longer) between the actual abnormal network activities and the ticket issued time.}, e.g., the time points near the ticketing time are highly likely to remain abnormal, but such ``closeness'' is hard to decide. (2)~\textit{imbalanced}, that is, the number of labeled abnormal samples is usually much smaller than the normal samples, making it challenging to learn representations from both positive and negative samples for predicting abnormal events accurately.

This paper proposes a machine learning approach, named SNetAD, that learns representative features from network operation time series by taking advantage of incomplete and imbalanced labels and uses the learned feature embeddings to predict abnormal events (e.g., packet loss). Specifically, SNetAD first constructs a graph to represent a given network where each node describes some attributes of an interface (e.g., in/out ucast packets) at a time point, and each edge is the connection between interfaces. Given a sequence of graphs (i.e., network data logs over a time period), SNetAD generates an embedding to represent network behaviors by jointly capturing local features within a temporal neighborhood and global features from neighboring nodes due to data transmission. 
During optimization, SNetAD encourages these learned embeddings to form two separable clusters representing normal and abnormal behaviors, respectively.\footnote{The definition of ``normal'' and ``anomalous'' could vary among applications. In this work, we assume there exists a feature space in which normal network activities are as far as possible from abnormal operations so that normal samples can be treated as one cluster and anomalies as another cluster.} SNetAD first leverages labeled samples to explicitly learn the cluster centers optimized by contrastive center loss. However, due to limited and imbalanced labeled samples, the learned centers might not be applicable to all samples, resulting in poor embedding quality. Thus, SNetAD updates the embeddings of all samples towards a high confidence clustering distribution to enforce cluster cleanness with a semi-supervised learning strategy. The goal is to improve the separability of the learned clusters representing normal and abnormal network behaviors. Finally, SNetAD utilizes a support vector classifier on these embeddings to predict if the given graph sequence of network operations would lead to an abnormal event.

In sum, the main contribution of this paper is a semi-supervised learning approach that learns feature embeddings by jointly capturing complex dependencies in multiple multivariate time series and enhances the separability of the embeddings for predicting abnormal events guided by the incomplete and imbalanced labeled data. We demonstrate the proposed network architecture for abnormal event prediction on large real-world network operation time-series data.

\section{Methodology}
\subsection{Problem Definition}

Given a network containing multiple interfaces with unknown connectivity, SNetAD first constructs a fully-connected graph, $\boldsymbol{G}=(\boldsymbol{V}, \boldsymbol{E})$, where each node $v \in \boldsymbol{V}$ represents an interface ($|\boldsymbol{V}|=n$) and the connection between any pair of nodes is initialized as one, meaning there can be data communication between interfaces. Let $\boldsymbol{G}^{(t)} \in \mathbb{R}^{n \times F}$ be the graph for network operation measurements at time~$t$, where~$n$ is the number of nodes and~$F$ is the number of attributes describing each node. Suppose the input is a sequence of graphs of length $T$, i.e., previous network activities from $t-T+1$ to $t$, SNetAD aims to learn a function $f$ that maps $T$ historical input signals to a binary signal $y \in \{0, 1\}$ indicating if there will be an abnormal event in the network at $t+1$.

\begin{displaymath}
[\boldsymbol{G}^{(t-T+1)}, \cdots, \boldsymbol{G}^{(t)}]\xrightarrow{\emph{f}} y
\end{displaymath}

\begin{figure}[htbp]
  \centering
  \includegraphics[width=\linewidth]{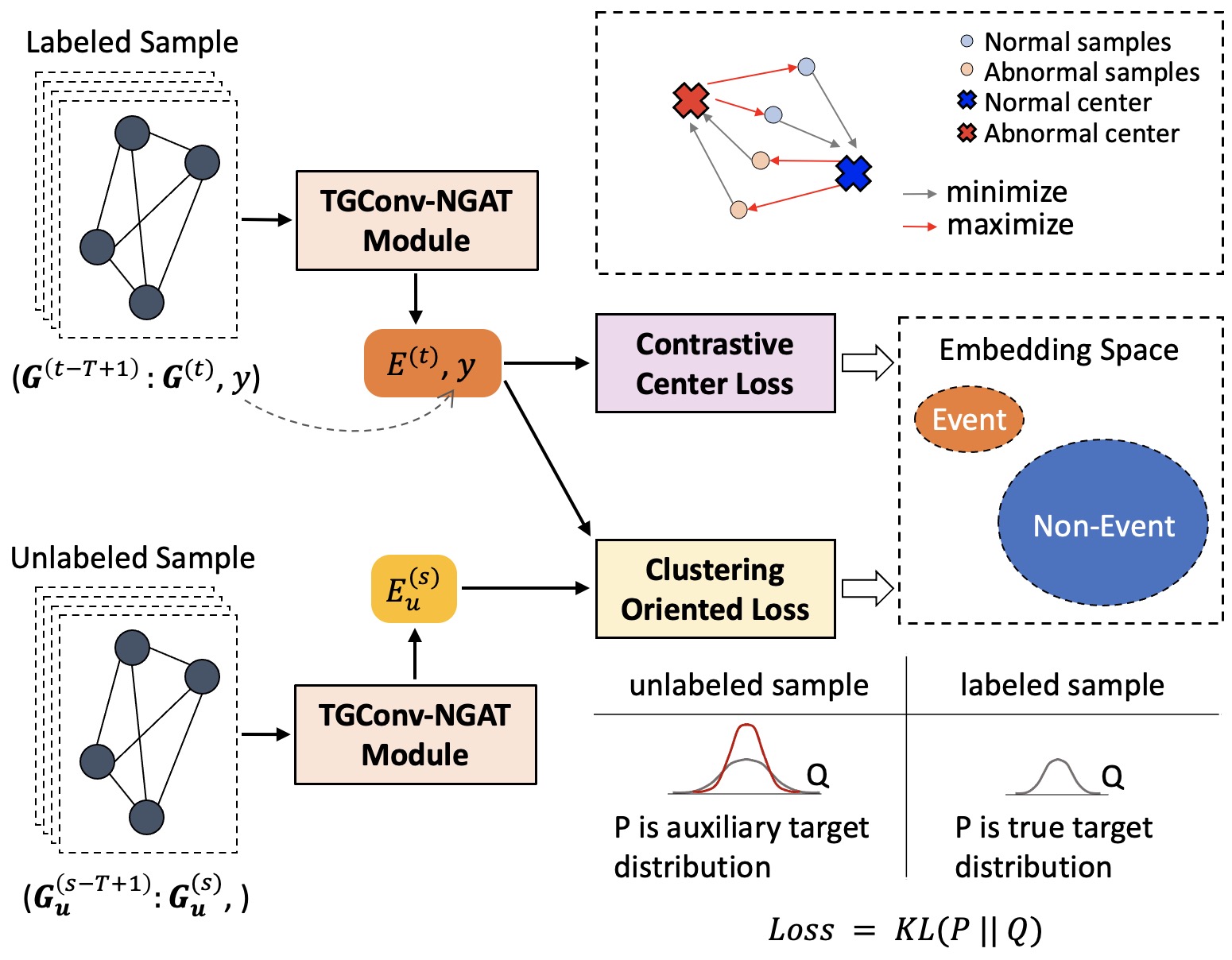}
  \caption{The network architecture and the training process of SNetAD}
  \label{figure: whole}  
\end{figure}

Figure~\ref{figure: whole} shows the proposed network architecture, SNetAD, for predicting abnormal events. Specifically, SNetAD first builds a TGConv-NGAT module to learn an initial feature embedding that represents network activities by jointly capturing dependencies across time points in the sequence and between nodes in the individual graph (Section~\ref{sec: stconvngat}). Then SNetAD refines the embeddings by (1) leveraging labeled normal and abnormal samples to form two separable clusters in a contrastive learning way, and (2) amplifying the “clustering cleanness” using labeled and unlabeled samples in a semi-supervised manner (Section~\ref{sec: cc}). Finally, our approach trains a support vector classifier with these learned embeddings for the prediction task (Section~\ref{sec: svc}).

\subsection{TGConv-NGAT Module} \label{sec: stconvngat}
\begin{figure}[htbp]
  \centering
  \includegraphics[width=\linewidth]{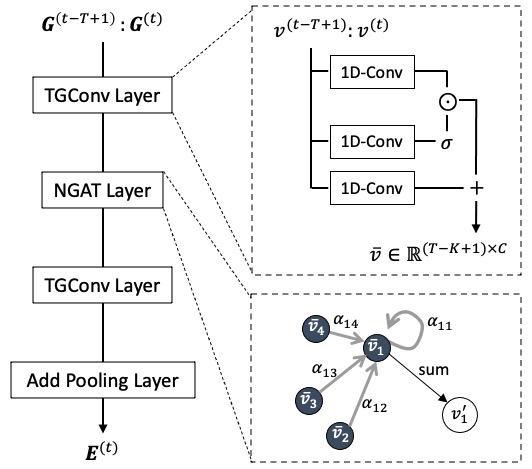}
  \caption{TGConv-NGAT module generates embeddings for the input graph sequences. The labeled and unlabeled samples share the same model weights.}
  \label{figure: stconvngat}  
\end{figure}

The paper proposes TGConv-NGAT module to generate feature embeddings from the input graph sequence (see the network architecture in Figure~\ref{figure: stconvngat}). In general, TGConv-NGAT module contains two temporal gated convolution (TGConv) layers to capture temporal dependencies across time, between which a normalized graph attention (NGAT) layer captures data transmission dependencies between nodes in the graph. The details of each component are described as follows. 

\subsubsection{TGConv Layer} Because the input graph sequence covers a wide time range (e.g., 6 hours) while abnormal network activities might happen within a local temporal neighborhood (e.g., only 30 minutes), this component aims to extract representative temporal features that lead to an abnormal event. 

Many existing work use recurrent neural networks (RNNs), e.g., long short-term memory (LSTM)~\cite{hochreiter1997long}, and Transformers~\cite{vaswani2017attention} to learn representations for sequential data by capturing temporal dependencies, but these models typically suffer from complex gating mechanisms, large memory requirement since every step relies on the previous steps, and time-consuming computation. In comparison, convolutional neural networks (CNNs) have the advantages of fast training and parallel computation~\cite{gehring2017convolutional} that can not only process the sequential data efficiently, but also explicitly capture temporal representative features within a local neighborhood that might contain abnormal activities.

Our approach adopts temporal gated-conv (TGConv) layers~\cite{gehring2017convolutional} that consist of 1-D convolution operations~\cite{dos2014deep} on the time axis to extract local temporal features for each node in the graph sequence. The gating mechanism in TGConv controls information passing through the gate. Specifically, the input to the 1-D convolution operations is a sequence of length $T$ with a channel size $F$ at each node, denoting as $v\in \mathbb{R}^{T \times F}$. With a kernel size~$K$ and a hidden size~$C$, three convolution kernels map the input to the output as $\mathbf{m}$, $\mathbf{n}$, and $\mathbf{b}$, respectively, where $\mathbf{m}$, $\mathbf{n}$, $\mathbf{b} \in \mathbb{R}^{(T-K+1) \times (C)}$. The gated-conv layer leverages gated linear units (GLU)~\cite{dauphin2017language} as a non-linear gating mechanism over $\mathbf{m}$ and $\mathbf{n}$:
\begin{equation}
    \text{GatedConv}(\mathbf{v}) = \mathbf{m} \odot \sigma(\mathbf{n}) + \mathbf{b}\ \in \mathbb{R}^{(T-K+1) \times (C)}
\end{equation}
where $\mathbf{m}$ and $\mathbf{n}$ are the input to the gates of GLU, $\mathbf{b}$ serves as a linear path, $\odot$ denotes the element-wise product, $\sigma$ is the sigmoid function, and $\sigma(\mathbf{n})$ controls which information in $\mathbf{m}$ are relevant to generating temporal features. Thus, the TGConv layer embeds each node $v$ as $(T-K+1)$ feature vectors of size $C$, where each vector contains the extracted temporal features for the corresponding kernel windows.

\subsubsection{NGAT layer} In addition to the temporal features learned from individual nodes, capturing dependencies between nodes is also essential for determining an abnormal event, e.g., packet losses during data transmission. Previous work in various domains apply graph convolution for modeling nodes' relationships, which require the network connectivity and careful design on the propagation functions~\cite{kipf2016semi, li2017diffusion, defferrard2016convolutional}. 

Since the graph topology that indicates the network connectivity could be unknown, our approach introduces a normalized graph attention (NGAT) layer to capture relationships between nodes by learning the propagation weight from node features in an arbitrary graph. Specifically, a graph attention (GAT) layer~\cite{velivckovic2017graph} computes the representation for each node by aggregating information from other neighboring nodes as:
\begin{equation}
    \text{GAT}(i) = \sigma \Big ( \sum_{k\in \mathcal{N}_i}\alpha_{ik}\mathbf{v_k} \Big )
\end{equation}
where node $i$ is the target node, $\mathbf{v_i} \in \mathbb{R}^{C}$ is the feature vector of node $i$, $\mathcal{N}_i$ is the neighborhood set of node $i$, $\sigma$ is the sigmoid function, $\alpha_{ik}$ is the attention score for node~$k$ to node~$i$. The attention score $\alpha_{ik}$ is defined by:
\begin{equation}
   \alpha_{ik}=\frac{\text{exp}(\sigma(\mathbf{w}^T \cdot (\mathbf{v_i}\ \Vert\ \mathbf{v_k}))}{\sum_{j \in \mathcal{N}_i} \text{exp}(\sigma(\mathbf{w}^T \cdot (\mathbf{v_i}\ \Vert\ \mathbf{v_j}))))}
\end{equation}
where the operation $\Vert$ is to concatenate two vectors, $\mathbf{w} \in \mathbb{R}^{2C}$ is a learnable weight vector, $\mathcal{N}_i$ is the neighborhood set of node $i$, $\sigma$ is the LeakyReLU function. $\alpha_{ik}$ serves as a learnable edge weight between two nodes and more important nodes receive higher attention during neighborhood aggregation.   

Our approach adds graph normalization (GraphNorm)~\cite{cai2021graphnorm} to the output of the GAT layer to accelerate the optimization process. Many existing work apply LayerNorm~\cite{ulyanov2016instance, xu2019understanding} after attention layers, e.g., Transformers~\cite{vaswani2017attention}), to brings distribution stability of the output and improve training efficiency. Unlike LayerNorm that performs normalization across neurons in a layer, GraphNorm normalizes all nodes for each individual graph with a learnable shift, defined as follows:
\begin{equation}
    \text{GraphNorm}(h_{i, j})=\gamma_j \cdot \frac{h_{i, j} - \alpha_j \mu_j}{\sigma_j} + \beta_j
\end{equation}
\begin{displaymath}
    \mu_j=\frac{\sum_{k=1}^n h_{k, j}}{n}, \  \sigma_j=\frac{\sum_{k=1}^n (h_{k, j} - \alpha_j \mu_j)}{n}
\end{displaymath}
where $h_{i, j}$ is the value of the $j$th feature dimension of $\text{GAT}(i)$ (the output of GAT layer for node $i$), $n$ is the number of nodes in the graph, $\alpha_j$ is the learnable parameter for the feature dimension $j$, which controls how much information to keep in the mean. $\gamma_j$ and $\beta_j$ are the affine parameters. 

\subsubsection{Putting everything together} Inspired by~\cite{yu2017spatio}, our approach adds another TGConv layer after the NGAT layer to fuse the features of node interactions on the time axis. This architecture can achieve fast node-state propagation through temporal layers and can be stacked multiple times. After the second TGConv layer, our approach concatenates the output to generate the node embeddings. Then our approach uses a global pooling layer to add node embeddings across the node dimension to generate the graph embedding. In this way, given [$\boldsymbol{G}^{(t-T+1)}, \cdots, \boldsymbol{G}^{(t)}] \in \mathbb{R}^{T \times n \times F}$, the TGConv-NGAT module generates the feature embedding $\boldsymbol{E}^{(t)}$ of length  ${(T-2(K-1)) \times C}$ that jointly captures the temporal dependencies and the relationships between nodes.

\subsection{Learning Clustering Structure} \label{sec: cc}
SNetAD assumes there exists a feature space in which normal data form a cluster, and abnormal data can form another cluster. This section introduces a semi-supervised strategy for refining the learned representation space to form the two separable clusters .

\subsubsection{Learning clustering structure with labeled samples in a contrastive way} SNetAD first randomly initializes the cluster centers representing normal and abnormal events, respectively. Then the labeled samples can explicitly guide the process of learning cluster centers as well as updating the feature embeddings by enforcing the embeddings of normal samples to be close to the normal center and far away from the abnormal center and vice versa. SNetAD uses the contrastive center loss that enhances intra-class compactness and inter-class separability simultaneously~\cite{qi2017contrastive}. Specifically, SNetAD penalizes the contrastive values between (1) the distance of a sample to its corresponding class center and (2) the distance of the sample to the non-corresponding class center,
\begin{equation}
    \mathcal{L}_{cc} = \frac{1}{2} \sum^m_{i=1} \frac{\lVert \mathbf{E}_i - \mathbf{C}_{y_i} \rVert^2_2}{\lVert \mathbf{E}_i - \mathbf{C}_{\bar{y}_i} \rVert^2_2 + \epsilon } 
\label{eq: cc}
\end{equation}
where $\mathcal{L}_{cc}$ denotes the contrastive center loss, $m$ is the number of labeled samples in a mini-batch, $\mathbf{E}_i$ is the embedding of the $i$-th training sample\footnote{The superscript $(t)$ is dropped for simplicity.}, $\mathbf{C}_{y_i}$ is the corresponding center of $\mathbf{E}_i$ and $\mathbf{C}_{\bar{y}_i}$ is the opposite one. When $\mathbf{C}_{y_i}$ is the normal center, $\mathbf{C}_{\bar{y}_i}$ represents the abnormal center, and vice versa. $\epsilon$~is a constant for preventing the denominator equal to 0 and set to ${1}\mathrm{e}{-6}$ empirically.

\subsubsection{Learning clustering structure with all samples in a semi-supervised way} 
Due to the limited and imbalanced labeled samples, the learned centers are easily guided by the given labeled samples and might fail during inference. To overcome the scarcity of labeled training samples and ensure the clusters are applicable to all samples, our approach introduces a clustering refinement process that iteratively minimizes the distance between the cluster assignment and some target cluster distribution using all samples in a semi-supervised manner. For the labeled samples, the target cluster distribution can be directly derived from the labels, while for the unlabeled samples, the target distribution is derived from high confidence predictions of the current cluster distribution. 

SNetAD first calculates the probability of assigning the feature embedding $\mathbf{E}_i$ to the cluster center $\mathbf{C}_0$ (i.e., normal center) and $\mathbf{C}_1$ (i.e., abnormal center) using inverse distance,
\begin{equation}
\begin{split}
    q_{i0} &= \frac{(1+\rVert \mathbf{E_i} - \mathbf{C}_0 \rVert^2_2) ^ {-1}}{(1+\rVert \mathbf{E_i} - \mathbf{C}_0 \rVert^2_2) ^ {-1} + (1+\rVert \mathbf{E_i} - \mathbf{C}_1 \rVert^2_2) ^ {-1}} \\
    q_{i1} &= \frac{(1+\rVert \mathbf{E_i} - \mathbf{C}_1 \rVert^2_2) ^ {-1}}{(1+\rVert \mathbf{E_i} - \mathbf{C}_0 \rVert^2_2) ^ {-1} + (1+\rVert \mathbf{E_i} - \mathbf{C}_1 \rVert^2_2) ^ {-1}}
\end{split}
\label{eq: Q1}
\end{equation}
where $q_{i0}$ and $q_{i1}$ are the probability of cluster assignment.

SNetAD then derives an auxiliary distribution made up of the high confidence assignments of the current cluster distribution . The main idea is to amplify the cluster assignment with high confidence to enhance the cluster purity. Let $p_{i0}$ and $p_{i1}$ be the auxiliary distribution. When i is a labeled normal sample: $p_{i0} = 1$ and $p_{i1} = 0$. When i is a labeled abnormal sample, $p_{i1} = 1$ and $p_{i0} = 0$. When i is an unlabeled sample, Equation~\ref{eq: P} shows the computation of auxiliary distribution following~\cite{xie2016unsupervised}:
\begin{equation}
\begin{split}
    & p_{i0} = \frac{q_{i0}^2 / \sum_{i'} q_{i'0}}{q_{i0}^2 / \sum_{i'} q_{i'0} + q_{i1}^2 / \sum_{i'} q_{i'1}} \\
    & p_{i1} = \frac{q_{i1}^2 / \sum_{i'} q_{i'1}}{q_{i0}^2 / \sum_{i'} q_{i'0} + q_{i1}^2 / \sum_{i'} q_{i'1}} \\
\end{split}
\label{eq: P}
\end{equation}
The second power of $q_{i0}$ and $q_{i1}$ places more weight on the instances near the center due to the higher assignment probability. The division of $\sum_{i'} q_{i'0}$ and $\sum_{i'} q_{i'1}$ normalizes based on the cluster size, making the model robust to biased classes. Note that the labeled samples use the true distribution as the auxiliary distribution and hence enable the training process in a semi-supervised way. 

To measure the distance between the distributions $p_{i0}$ and $q_{i0}$ (also $p_{i1}$ and $q_{i1}$), SNetAD computes the KL divergence as the clustering oriented loss,
\begin{equation}
   \mathcal{L_{KL}} = \sum_i p_{i0} \log \frac{p_{i0}}{q_{i0}} + p_{i1} \log \frac{p_{i1}}{q_{i1}}
\label{eq: Q2}
\end{equation}

Therefore, the training process of SNetAD is guided by the sum of the contrastive-center loss and the clustering oriented loss with a hyper-parameter $\lambda$,
\begin{equation}
   \mathcal{L} = \mathcal{L}_{cc} + \lambda \times \mathcal{L}_{KL}
\label{eq: Q3}
\end{equation}


\subsection{Predicting Abnormal Events Using SVC}\label{sec: svc} 
\begin{figure}[htbp]
  \centering
  \includegraphics[width=\linewidth]{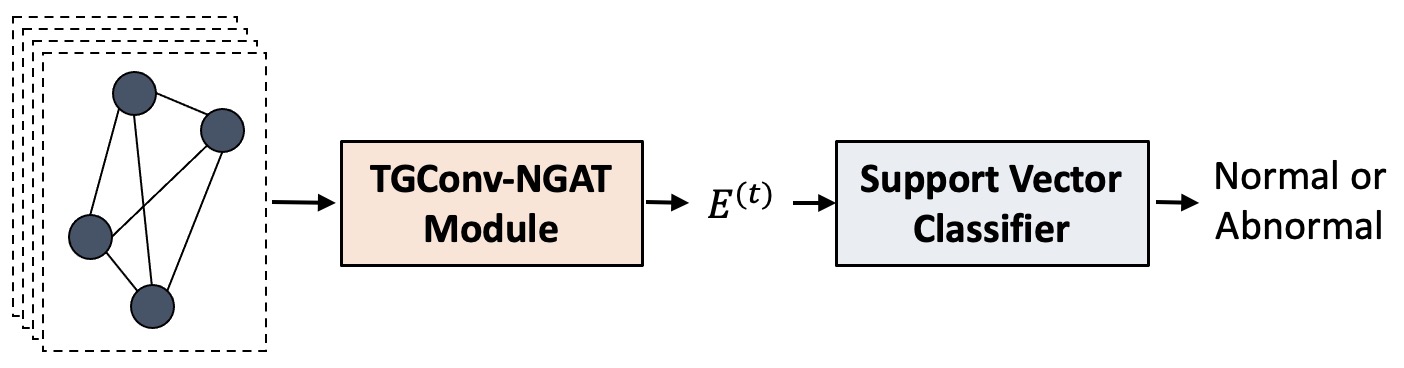}
  \caption{The inference process takes a graph sequence to generate a feature embedding and classifies it as normal or abnormal.}
  \label{figure: inference}  
\end{figure}

After using the TGConv-NGAT to learn feature embeddings from the graph sequence and the semi-supervised loss to improve the separability of the embeddings, SNetAD takes the embeddings and labels to train a support vector classifier (SVC) to predict normal and abnormal events. SNetAD does not train in an end-to-end manner, e.g., adding a fully-connected (FC) layer for classification, because SVC can identify a hyperplane for the best possible separations and usually works better than FC if the input is already separable in some space~\cite{sahay2015svm}. Figure~\ref{figure: inference} shows the inference process. Given a graph sequence representing network operations, the TGConv-NGAT module generates the feature embedding, and the SVC takes the embedding to predict if the input would lead to an abnormal event. 

\section{Experiments and Results}
We implemented the proposed architecture with PyTorch~1.8 and PyTorch Geometric framework~\cite{Fey/Lenssen/2019}. We conducted the experiments on a GPU server with one physical core and 16GB memory. Our code is available on \url{https://github.com/knowledge-computing/semisup-abnormal-event-prediction}.

\subsection{Datasets} 
We conducted the experiments with the network operation data from NTT Global Networks (NTT-GN). The dataset contains more than 150 million data logs from network interfaces (e.g., WAN routers and LAN switches) that NTT-GN collects, monitors, and manages. The data logs cover the time range from 2020-05-12 to 2020-06-12, with a time interval of 5 minutes. Figure~\ref{figure: network_def} shows an example of the network structure. In this work, we address a practical case where there could be dependencies between any interface pairs in a network (e.g., data transmissions), and the network topology is unavailable. The graph construction can be modified if prior knowledge of the network topology is available.  

In addition to the network operation data, NTT-GN provides a packet loss event dataset that reports the event's ticket opening time. We assume that some abnormal network activities before the ticketing time cause the events, where the gap between the actual event happening time and the ticket issued time is within 6 hours, and the event would be resolved in 6 hours.\footnote{Because the event starting time and resolving time are unavailable, we confirm with NTT-GN network experts about the decision of using 6 hours as the window size for the events.} We construct samples with graph sequence of window size 72 (i.e., 6 hours), and SNetAD aims to capture anomalies by looking at the dynamics of the graph sequence. 

\begin{figure}[htbp]
\centering
    \includegraphics[width=\linewidth]{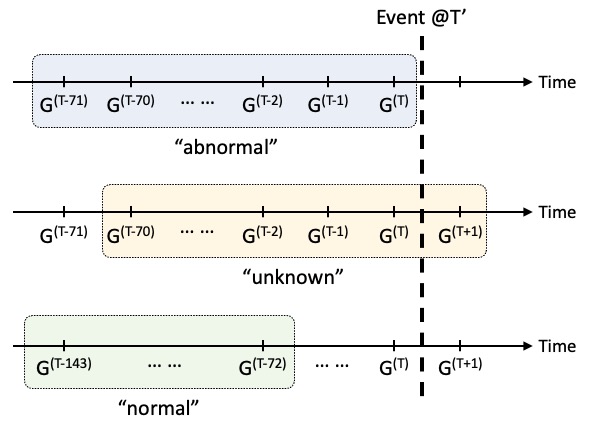}
\caption{Examples of labeling ``abnormal'', ``unknown'', and ``normal'' to samples given an event's ticketing time}
\label{figure: label_construction}  
\end{figure}

Figure~\ref{figure: label_construction} shows an example of labeling a graph sequence. If there is an event reported at $T'$ and the closest graph before that is $\mathbf{G}^{(T)}$, we assigned an ``abnormal'' label to the sample of graph sequence $[\mathbf{G}^{(T-71)}, ..., \mathbf{G}^{(T)}]$ (e.g., blue box), denoting that these graphs contain some abnormal network activities directly leading to an event. If there is an event reported within 6 hours before and after a graph sequence, we marked the sample as ``unknown'' (e.g., yellow box), representing unlabeled data. The rest samples that do not overlap with any event within 6 hours are labeled as ``normal'' (e.g., green box). These labeled and unlabeled samples will jointly improve feature embeddings and clusters in a semi-supervised setting for abnormal event prediction.

To evaluate the proposed approach, we selected the available networks with at least 50 events. Table~\ref{table: dataset summary} shows the summary of the dataset. A larger number of events results in more ``unknown'' labels, i.e., the ratio of labeled samples to unlabeled samples becomes lower (i.e., label incompleteness). The number of normal samples is much larger than the abnormal samples (i.e., label imbalance). 

\begin{table*}[htbp]
\centering
\caption{Summary of the dataset, ordered by the number of events}
\renewcommand\arraystretch{1.3}
\label{table: dataset summary}
\begin{tabular}{ >{\centering\arraybackslash}p{1.8cm} || c | c | c | c | c | c} 
\hline 
    Network ID & \#Nodes & \#Abnormal & \#Labeled & \#Unlabeled & \#Labeled : \#Unlabeled & \#Normal : \#Abnormal  \\ \hline \hline
    1NTT18635 & 143 & 189 & 1648 & 7786 & 0.175 : 0.825 & 0.885 : 0.115 \\ \hline
    1NTT19478 & 120 & 147 & 2044 & 7388 & 0.217 : 0.783 & 0.928 : 0.072 \\ \hline
    1NTT19535 & 84  & 142 & 2327 & 7105 & 0.247 : 0.753 & 0.939 : 0.061 \\ \hline
    1NTT67246 & 36  & 134 & 2608 & 6824 & 0.277 : 0.723 & 0.949 : 0.051 \\ \hline
    1NTT18927 & 53  & 125 & 4070 & 5362 & 0.432 : 0.568 & 0.969 : 0.031 \\ \hline
    1NTT22885 & 91  & 91  & 2633 & 6799 & 0.279 : 0.721 & 0.965 : 0.035 \\ \hline
    1NTT48337 & 89  & 83  & 2857 & 6575 & 0.303 : 0.697 & 0.971 : 0.029 \\ \hline
    1NTT19213 & 119 & 78  & 3396 & 6037 & 0.360 : 0.640 & 0.977 : 0.023 \\ \hline
    143XB8CT  & 97  & 78  & 4052 & 5380 & 0.430 : 0.570 & 0.981 : 0.019 \\ \hline
    1NTT30990 & 20  & 71  & 4603 & 4829 & 0.488 : 0.512 & 0.985 : 0.015 \\ \hline
\end{tabular}
\end{table*}

\subsection{Experiment Settings} 
\subsubsection{Training Settings} We split the samples into training, validation, and testing sets, ensuring each set contains both normal and abnormal samples. We normalize the samples with the maximum and minimum values from the training data. We set the batch size as 16, the hidden state of temporal gated convolution as 32, the kernel size of 1D convolution as 12 (i.e., 1-hour window size), and the output feature embedding size as 256. The initial learning rate was 0.001, with early stopping on the validation dataset. The hyper-parameters were chosen by grid search on the validation set, and we reported the results with the set of hyper-parameters yielding the best average performance, that is $\lambda=0.1$. 

\subsubsection{Metrics} We report the precision, recall, and F1 score for ``abnormal'' since our goal is to predict abnormal events. The higher the values the better the performance.\footnote{We use label 1 to represent ``abnormal'' and label 0 to represent ``normal'' for simplicity.}
\begin{equation*}
    \text{precision(label=1)} = \frac{\text{TP(label=1)}}{\text{TP(label=1) + FP(label=1)}}
\end{equation*}
\begin{equation*}
\text{recall(label=1)} = \frac{\text{TP(label=1)}}{\text{TP(label=1) + FN(label=1)}}
\end{equation*}
\begin{equation*}
\text{F1} = 2 \times \frac{\text{precision(label=1)} \times \text{recall(label=1)}}{\text{precision(label=1) + recall(label=1)}}
\end{equation*}

\subsubsection{Baselines} We compare SNetAD to the following state-of-the-art baselines:
\begin{itemize}
    \item Support Vector Classifier (SVC): SVC is a classic supervised learning method. We convert the problem to a binary classification task for SVC. We flatten the input graph sequence into a large vector and use labeled samples to train an SVC for predicting anomalous events.  
    \item One class SVM (OCSVM)\cite{wang2004anomaly}: OCSVM is a typical unsupervised learning approach for anomaly detection in machine learning. We flatten the input in the same way as SVC and use all samples (labeled and unlabeled) to fit the model.
    \item MTAD-GAT~\cite{zhao2020multivariate}: This is an unsupervised anomaly detection approach that learns feature embeddings by using two attention layers to capture relationships between nodes and across time points, respectively, and then computes reconstruction loss and forecasting loss simultaneously as the inference score for anomaly detection. 
    \item Graph Attention Encoder (GAT-Semi)~\cite{velivckovic2017graph}: We fuse the features along the time dimension for each node and use GAT only to generate feature embeddings. This method aims to demonstrate the necessity of capturing local temporal dependencies. The clustering architecture remains the same as SNetAD.
    \item Graph Attention LSTM (GAT-LSTM-Semi)~\cite{wu2018graph}: Instead of TGConv layer, GAT-LSTM uses LSTM to capture the interactions across time points. The clustering architecture remains the same as SNetAD.
    \item DeepSAD~\cite{ruff2019deep}: This is a semi-supervised learning approach for anomaly detection that learns a cluster center to represent normal cases. The original domain is computer vision. To extend the model on our dataset, we apply the TGConv-NGAT module as the encoder.
    \item Model variants: (1) TGConv-NGAT + only contrastive center loss (only-CC), which is trained only with labeled samples; (2) TGConv-NGAT + only KL divergence loss (only-KL); (3) We construct a correlation graph as the network connectivity following~\cite{deng2021graph}, which computes pairwise Pearson correlation between nodes and assumes two nodes are connected only if the correlation is above the mean plus one std (Corr-Graph).
\end{itemize}


\subsection{Model Performance.}
\begin{figure*}[htbp]
\centering
    \includegraphics[width=\linewidth]{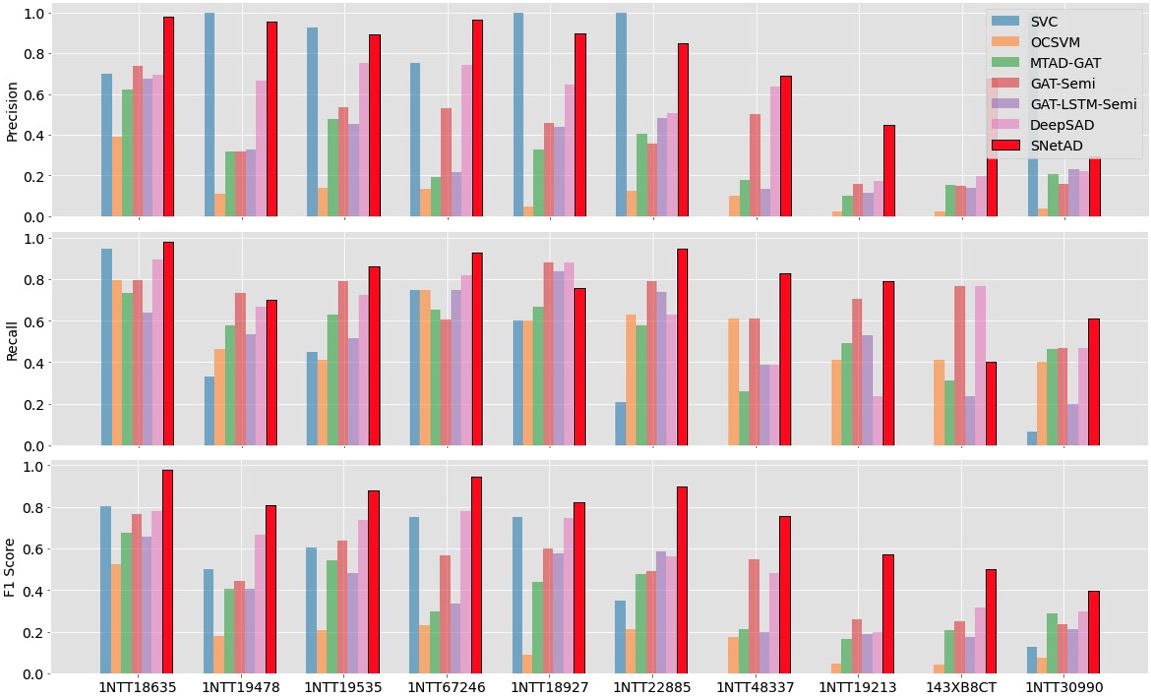}
\caption{The precision, recall, and F1 scores for baselines and SNetAD (red).}
\label{figure: f1}  
\end{figure*}

Figure~\ref{figure: f1} shows the precision, recall, and F1 scores for the baselines and SNetAD. Our approach outperforms baseline methods in all networks. SNetAD achieves an F1 score above 0.7 in seven out of ten networks. The F1 scores generally decrease when there is a fewer number of labeled abnormal events, which might be because limited abnormal samples are not enough to learn a representative cluster center and lead to failure in some testing cases.

Specifically, SNetAD achieves more than 50\% improvement on average in F1 scores compared to the supervised approach, SVC. SVC results in high precision but low recall for networks with relatively large abnormal labels. Due to imbalanced labeled samples, the classifier tends to predict normal, and anomalies are misclassified during testing. Also, SVC takes the input graph sequence as a flattened vector instead of a high-level feature representation, making it hard to find a proper hyperplane for classification. 

Unsupervised learning approaches, OCSVM and MTAD-GAT, typically have high recall but low precision, resulting in poorer performance than SNetSD. MTAD-GAT performs better than OCSVM by carefully designing the neural network architecture to capture representative features from the input graph sequence and applying two scoring strategies to make robust anomaly inference. However, assuming rare samples as anomalies is applicable in practice. SNetSD adds labeled samples during training that guide the learning process to generate representative embeddings.

\begin{figure}[htbp]
\centering
    \includegraphics[width=\linewidth]{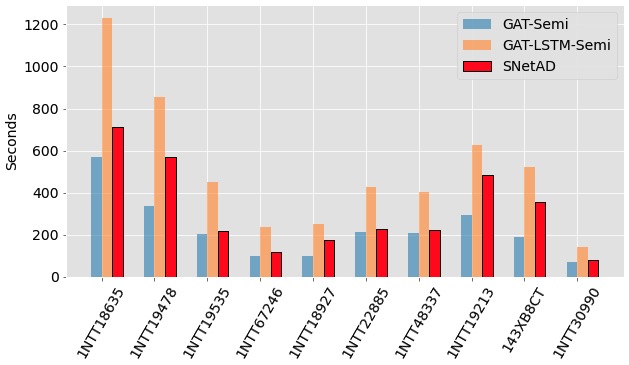}
\caption{The training time comparison for running approximate 9734 samples with batch size 16 in one epoch}
\label{figure: speed}  
\end{figure}

In terms of encoding the input graph sequence, SNetAD with the TGConv-NGAT module shows advantage over GAT-Semi and GAT-LSTM-Semi. First, GAT does not explicitly consider temporal dependencies but fuses features on the time dimension and expects the model to extract useful information from these values for prediction directly. However, the temporal order matters for the event prediction, e.g., the position of a sudden increment might result in different network behaviors. Our approach leverages the TGConv layer to summarize local temporal features, aiming to extract effective temporal patterns leading to an event. 
Second, although GAT-LSTM can capture dependencies between nodes and across time points in multiple time series~\cite{wu2018graph}, the performance is poorer than SNetSD because some effective patterns (or features) at the early stage of the input sequence might be easily weakened or neglected during the sequential training of LSTM. However, in practice, only some subsequence(s) from the input graph sequence would cause an event. Since the impacting subsequence(s) on the event are unknown (i.e., the leading time to an event could be varying), the sequence length (i.e., six hours in the experiment) is chosen to cover as many time points as possible. In comparison, the TGConv layer focuses on manipulating temporal interactions on every subsequence, which empirically fulfills our assumption and achieves better performance on real datasets. Moreover, Figure~\ref{figure: speed} shows that SNetAD requires much less training time than LSTM based approach.

SNetAD outperforms other semi-supervised approaches for anomaly detection, such as DeepSAD. The loss function in DeepSAD enforces labeled normal samples to be close to the predefined center and abnormal samples to be as far as possible. For the unlabeled samples, DeepSAD has a strong assumption that the majority of data samples are normal and treats all unlabeled samples as normal in the loss. However, such an assumption might not hold in practice since one event can last for some period (i.e., the unlabeled data might contain abnormal network activities), which might confuse the model to learn effective representations for unlabeled samples and result in poor performance. 



\begin{figure}[htbp]
\centering
    \subfigure[Network ID: 1NTT19535]{
        \begin{minipage}[t]{\linewidth}
            \centering
            \includegraphics[width=\linewidth]{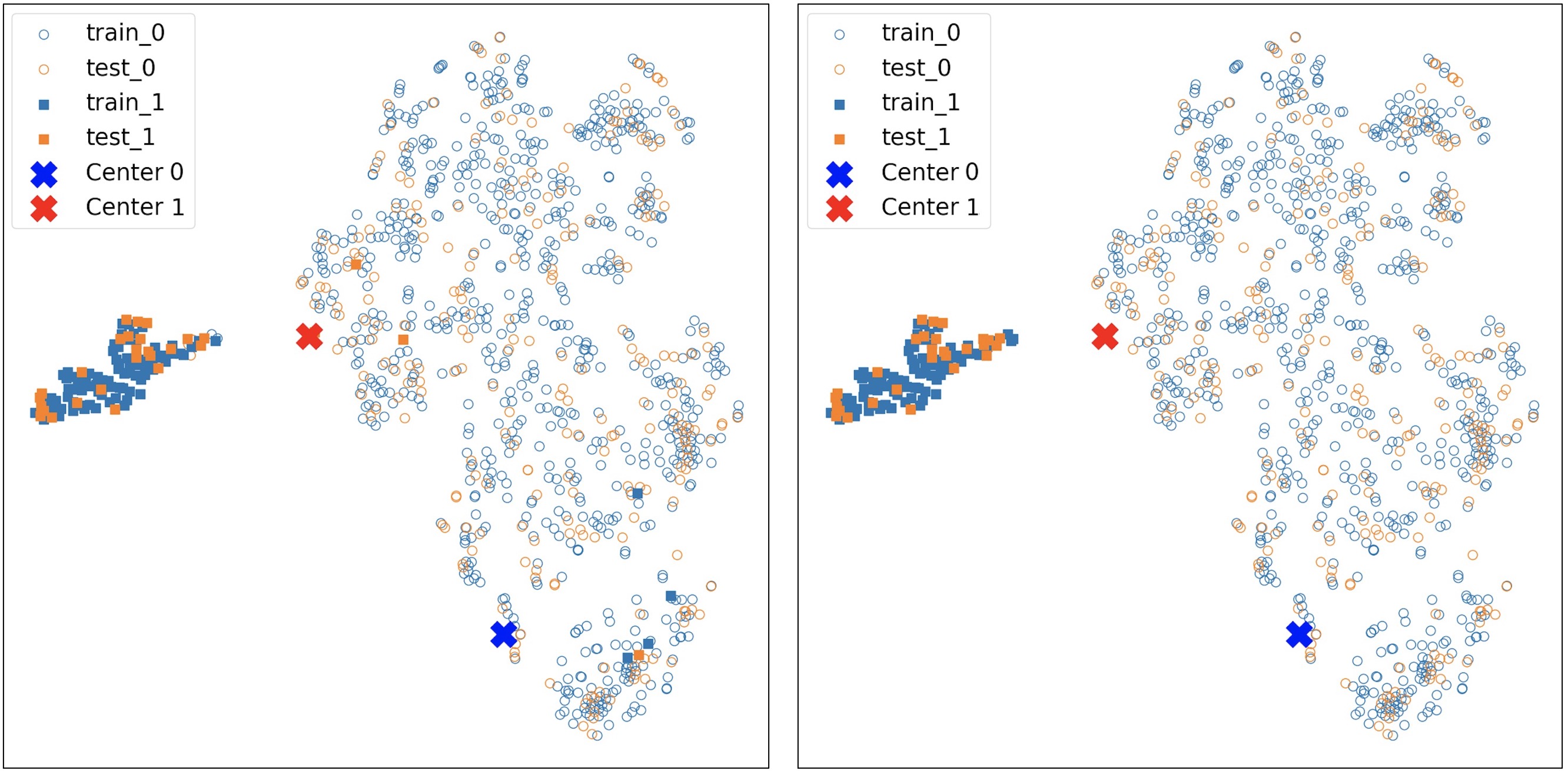}
            \label{fig: viz21}
        \end{minipage}
        }
    \subfigure[Network ID: 1NTT48337]{
        \begin{minipage}[t]{\linewidth}
            \centering
            \includegraphics[width=\linewidth]{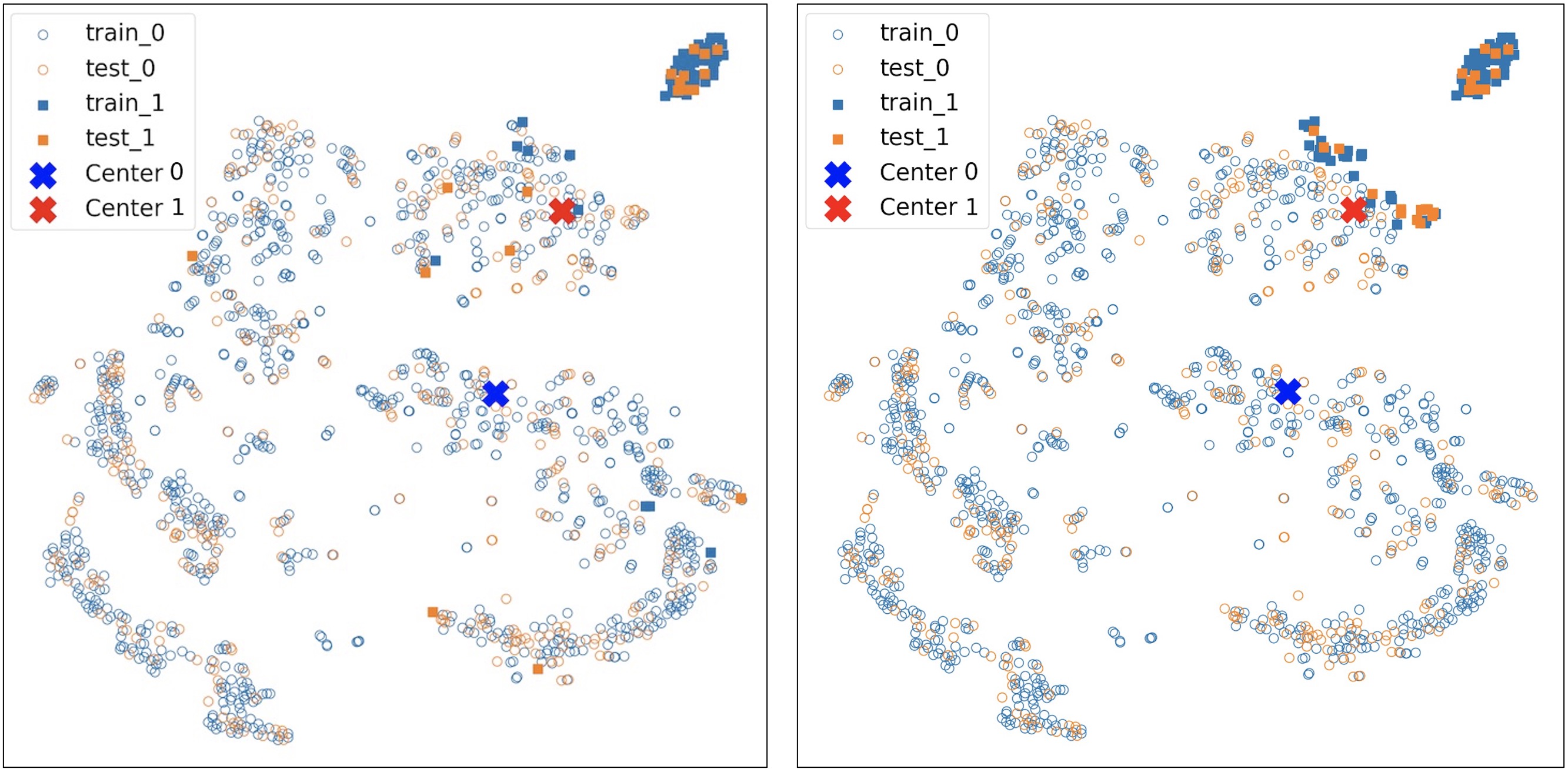}
            \label{fig: viz22}
        \end{minipage}
        }
\caption{Two examples of learned embedding space in 2D using TSNE. The left side shows the embeddings with true labels, and the right shows the prediction results. The hollow dots are for normal events, while the solid ones are for abnormal events.}
\label{figure: res2d}  
\end{figure}

We qualitatively show the abnormal event prediction performance by projecting the feature embeddings and cluster centers on a 2D space using TSNE~\cite{van2008visualizing}.\footnote{\url{https://scikit-learn.org/stable/modules/generated/sklearn.manifold.TSNE.html}} Figure~\ref{figure: res2d} shows the example plots of embeddings with the labels: the network ``1NTT19535'' has relatively high F1 score while ``1NTT48337'' is relatively low among the cases. We use blue and red crosses as the normal and abnormal cluster centers. The small blue dots represent training samples, and the orange ones are testing samples. The hollow circles are for normal events, while the solid squares are for abnormal events. The left figures show the embeddings for the true labels, and the right figures are for the prediction results. We observe that SNetAD learns a separable embedding space, i.e., normal samples (hollow circles) are typically clustered and far away from abnormal samples that form another cluster (solid squares). The network ``1NTT19535'' has few abnormal cases misclassified where the embeddings are blended in the normal cluster, which results in a good performance. The network ``1NTT48337'' has some abnormal embeddings far from the right corner cluster and scattered over the space, making the SVC challenging to separate them (i.e.,  predicting the samples above the red cross as abnormal) and resulting in relatively low performance. This might be because there are too few abnormal samples of these types of network activities in the training data to guide the network to extract representative abnormal patterns for them. Besides, we observe that within the normal cluster, the embeddings tend to form multiple sub-clusters (i.e., some embeddings are relatively closer to each other than others), indicating that various types of normal network activities could exist. However, SNetAD encourages all normal embeddings that potentially form multiple clusters to be far away from the abnormal cluster, so the embedding space still has good separability for robust prediction.


\subsection{Ablation Study.} 
\subsubsection{Effect of Loss Function} We continue to examine the effect of the losses. Table~\ref{table: loss} shows that SNetAD achieves 12\% improvement in F1 scores on average compared to only using the contrastive center loss (Only-CC) and 14\% improvement compared to only using the KL loss (Only-KL). The Only-CC models trained on limited labeled samples generally have high precision and low recall, which work well for the networks with large abnormal samples (e.g., more than 120 events in ``1NTT67246'' and ``1NTT18635''). However, the performance drops in the cases with only a few abnormal samples. In comparison, SNetAD trains on labeled and unlabeled samples that encourage the learned clustering structure to be applicable to all data samples, resulting in better performance, e.g., 48\% improvement in F1 score for ``1NTT30990'' and 35\% for ``1NTT19213''. In the other case, the Only-KL models trained with all samples do not explicitly optimize the centers, while SNetAD makes the best use of labeled samples to learn cluster centers, which improves the separability of the embeddings.

\subsubsection{Effect of Graph Construction} Some approaches add prior knowledge about the graph connectivity estimated from data. We follow ~\cite{deng2021graph} to construct graphs by computing pairwise Pearson correlation between nodes, where two nodes have a connection if their correlation is high; otherwise, there is no connection in between. Table~\ref{table: loss} shows the results comparison between using fully-connected graph (SNetAD) and correlation graph (Corr-Graph). The Corr-Graph models generally perform worse than SNetAD, which indicates that the node correlations can not effectively describe the interactions between interfaces. Since the network connectivity is unknown, SNetAD assumes that data communication (direct and indirect) could happen between any interface pairs, and using the graph attention module to dynamically capture nodes' relationships is flexible and achieves the best performance.


\begin{table*}[htpb]
\centering
\renewcommand\arraystretch{1.3}
\caption{Precision (P), Recall (R), and F1 comparison between SNetAD (the proposed approach), SNetAD with Only-CC and Only-KL (ablation study on the losses), and SNetAD with Corr-Graph (ablation study on the graph construction). \\ Models with the highest F1 scores are bold.}
\label{table: loss}
\begin{tabular}{ >{\centering\arraybackslash}p{1.8cm} || >{\raggedleft\arraybackslash}p{0.8cm} | >{\raggedleft\arraybackslash}p{0.8cm} | >{\raggedleft\arraybackslash}p{0.8cm} | >{\raggedleft\arraybackslash}p{0.8cm} | >{\raggedleft\arraybackslash}p{0.8cm} | >{\raggedleft\arraybackslash}p{0.8cm} | >{\raggedleft\arraybackslash}p{0.8cm} | >{\raggedleft\arraybackslash}p{0.8cm} | >{\raggedleft\arraybackslash}p{0.8cm} | >{\raggedleft\arraybackslash}p{0.8cm} | >{\raggedleft\arraybackslash}p{0.8cm} | >{\raggedleft\arraybackslash}p{0.8cm}} 
\hline
    & \multicolumn{3}{c|}{Only-CC} & \multicolumn{3}{c|}{Only-KL} & \multicolumn{3}{c|}{Corr-Graph} & \multicolumn{3}{c}{SNetAD} \\ \hline \hline
    Network ID & P & R & F1 & P & R & F1 & P & R & F1 & P & R & F1 \\ \hline 
    1NTT18635 & 0.947 & 0.923 & 0.935          & 0.950 & 0.974 & 0.962 & 0.617 & 0.860 & 0.719 & 0.978 & 0.982 & \textbf{0.980}\\
    1NTT19478 & 0.846 & 0.733 & 0.786          & 0.897 & 0.693 & 0.782 & 0.927 & 0.577 & 0.711 & 0.955 & 0.700 & \textbf{0.808}\\
    1NTT19535 & 0.913 & 0.724 & 0.808          & 0.876 & 0.692 & 0.773 & 0.775 & 0.701 & 0.736 & 0.893 & 0.862 & \textbf{0.877}\\ 
    1NTT67246 & 0.946 & 0.960 & \textbf{0.955} & 0.963 & 0.857 & 0.889 & 0.850 & 0.769 & 0.807 & 0.923 & 0.929 & 0.946\\ 
    1NTT18927 & 0.864 & 0.760 & 0.809          & 0.735 & 0.700 & 0.717 & 0.714 & 0.746 & 0.730 & 0.898 & 0.760 & \textbf{0.823}\\
    1NTT22885 & 0.875 & 0.730 & 0.800          & 0.810 & 0.895 & 0.850 & 0.553 & 0.791 & 0.651 & 0.850 & 0.949 & \textbf{0.897}\\ 
    1NTT48337 & 0.682 & 0.833 & \textbf{0.757} & 0.629 & 0.722 & 0.672 & 0.470 & 0.614 & 0.532 & 0.691 & 0.829 & 0.754\\
    1NTT19213 & 0.333 & 0.576 & 0.422          & 0.303 & 0.647 & 0.413 & 0.355 & 0.592 & 0.444 & 0.450 & 0.789 & \textbf{0.573}\\
    143XB8CT  & 0.667 & 0.353 & 0.462          & 0.621 & 0.373 & 0.466 & 0.495 & 0.366 & 0.421 & 0.674 & 0.400 & \textbf{0.502}\\ 
    1NTT30990 & 0.267 & 0.267 & 0.267          & 0.254 & 0.333 & 0.288 & 0.184 & 0.477 & 0.266 & 0.294 & 0.611 & \textbf{0.397}\\ \hline
\end{tabular}
\end{table*}

\section{Related Work}
Many existing approaches for anomaly detection with \textit{multiple multivariate time-series data} propose to learn representations that enable accurate anomaly prediction by capturing relationships across time points (e.g., temporal dependencies) and between time series (e.g., data transmissions). RNNs~\cite{malhotra2016lstm, hundman2018detecting} and Transformers~\cite{cai2020traffic, tuli2022tranad} can model temporal dependencies in time-series data, but they typically suffer from time-consuming training process and large memory requirement~\cite{yu2017spatio}. SNetAD uses the gated convolution to efficiently and effectively extract local temporal features that possibly lead to an abnormal event. Besides, existing work use feature fusion~\cite{malhotra2016lstm, li2019mad}, graph convolution~\cite{kipf2016semi, chen2022deep}, and graph attention~\cite{zhou2022hybrid, fan2020anomalydae} to model relationships between time series. SNetAD adopts graph attention with graph normalization to dynamically learn the weight between nodes from data since the network topology is unknown.

Most anomaly detection methods are unsupervised~\cite{munir2018deepant, deng2021graph}. Hundman et al.~\cite{hundman2018detecting} propose to use LSTM to predict spacecraft telemetry and detect outliers in an unsupervised manner. Zhao et al.~\cite{zhao2020multivariate} propose to combine a reconstruction-based model and a forecasting-based model to obtain some inference scores for anomalies. However, unsupervised approaches usually make a strong assumption that infrequent behavior is anomalous and usually requires some scores for inference (e.g., see \cite{ruff2018deep}). Although there are some strategies, e.g., Peak Over Threshold (POT)~\cite{siffer2017anomaly}, that automatically determine the threshold during inference, the performance is still not competitive to supervised methods when labels are available~\cite{liu2015opprentice, shipmon2017time}. Supervised approaches can explicitly learn normal or abnormal patterns from labels yet requires a prohibitive large number of labeled examples~\cite{blazquez2021review}. 

Semi-supervised methods overcome these limitations. Ruff et al.~\cite{ruff2019deep} demonstrate that adding a few labeled samples in a semi-supervised manner would guide the model to achieve better performance than unsupervised learning methods for anomaly detection. Some approaches optimize the learning process with a reconstruction loss for unlabeled samples plus a prediction loss for labeled samples (e.g., cross-entropy loss or focal loss for the imbalanced label situation)~\cite{xiao2020semi, idhammad2018semi, camacho2019semi}. However, these methods still require some threshold for inference and do not fully distinguish normal and abnormal events. In comparison, SNetAD explicitly learns a separable embedding space for normal and abnormal samples. Alsulami et al.~\cite{alsulami2022toward} presents a novel automatic labeling algorithm based on self-augmentation and ensemble classification strategies. However, label augmentation for anomalies might be misleading due to limited abnormal labels. Vercruyssen et al.~\cite{vercruyssen2018semi} propose to first uses a clustering-based approach to define typical normal behavior and identify anomalies. Then the approach actively adds labels from domain experts to guide the clustering approach in a semi-supervised manner. However, the performance would highly rely on the initialized clustering result. Instead, our approach gradually updates the embedding space from labeled and unlabeled samples to improve the ``cleanness'' of clusters in a self-learning way.

\section{Conclusion, Limitations, and Future Work}
The paper presents a semi-supervised learning approach, SNetAD, for predicting abnormal events from multiple multivariate time-series data. SNetAD learns a separable embedding space for normal and abnormal activities by encouraging embeddings to form two clusters and improving the separability of embeddings in a semi-supervised manner. This work has limitations in interpretability, e.g., determining the devices or interfaces and their activities leading to an event. Also, graph attention might be hard to scale when the network contains thousands of interfaces. We plan to explore the methods for detecting the event's root cause and extend SNetAD to handle extremely large networks. Another direction is to take advantage of other networks in a transfer learning way when the labeled data is scarce.

\section{Acknowledgement}
This material is based upon work supported in part by the NTT Global Networks and NVIDIA Corporation.

\bibliographystyle{IEEEtran}
\bibliography{references}

\end{document}